\documentclass[letterpaper, 10 pt, conference]{ieeeconf}  

\IEEEoverridecommandlockouts                              

\usepackage{cite}
\usepackage{amsmath,amssymb,amsfonts}
\usepackage{algorithmic}
\usepackage{graphicx}
\usepackage{textcomp}
\usepackage{xcolor}
\usepackage{multirow}
\usepackage{subcaption}
\usepackage{makecell}
\usepackage{hyperref}
\usepackage{booktabs}
\usepackage{placeins}

\def\BibTeX{{\rm B\kern-.05em{\sc i\kern-.025em b}\kern-.08em
    T\kern-.1667em\lower.7ex\hbox{E}\kern-.125emX}}

\title{\LARGE \bf
Taming Perception Jitter: Uncertainty-Aware LiDAR Object Detection for Reliable Motion Classification
}
\author{Cornelius Schröder$^{1}$, Žygimantas Marcinkus and Markus Lienkamp$^{1}$
\thanks{This research was funded by the Bavarian Research Foundation.}
\thanks{$^{1}$ Institute for Automotive Engineering,}
\thanks{\hspace{0.25cm}Munich Institute of Robotics and Machine Intelligence,}
\thanks{\hspace{2.5mm}School of Engineering and Design,}
\thanks{\hspace{2.5mm}Technical University of Munich}
\thanks{{\tt\small cornelius.schroeder@tum.de}}
}

\begin{document}

\maketitle
\thispagestyle{empty}
\pagestyle{empty}

\begin{abstract}
Reliable motion classification is critical for autonomous driving, as false dynamic predictions of static objects can cascade into unnecessary planner interventions. Unstable bounding box predictions can lead to spurious velocity estimates in tracking and falsely predicted trajectories. We present a deployment-friendly mitigation strategy that augments a 3D object detector with aleatoric uncertainty estimates and applies a two-sample z-test over short observation windows to separate true motion from jitter. Integrated into Autoware with minimal changes, the approach reuses existing data association for minimal compute overhead. Empirical results show parity with velocity thresholding on nuScenes, but substantially fewer false dynamic predictions and unnecessary stops in real-world test drives, explained by the presence of an intermediate “jitter band” in the recorded data that speed-only rules misclassify. This demonstrates that uncertainty-aware detection and lightweight statistical testing can deliver practical performance gains for autonomous driving in noisier real-world settings.
\end{abstract}

\section{Introduction}
Deploying an autonomous vehicle on public roads requires all perception and decision-making modules -- object detection, tracking, motion prediction, and trajectory planning -- to operate concurrently and interact reliably. In practice, these modules are coupled: imprecision in object detection propagates to tracking and prediction, creating new failure modes for planning. Performance further degrades when the sensor suite or operating environment differs from the available large-scale training data (e.g., nuScenes, Waymo), introducing domain gaps that are costly to overcome due to the need for large, matching datasets.

\begin{figure}[h]
    \centering
    \includegraphics[width=0.95\linewidth]{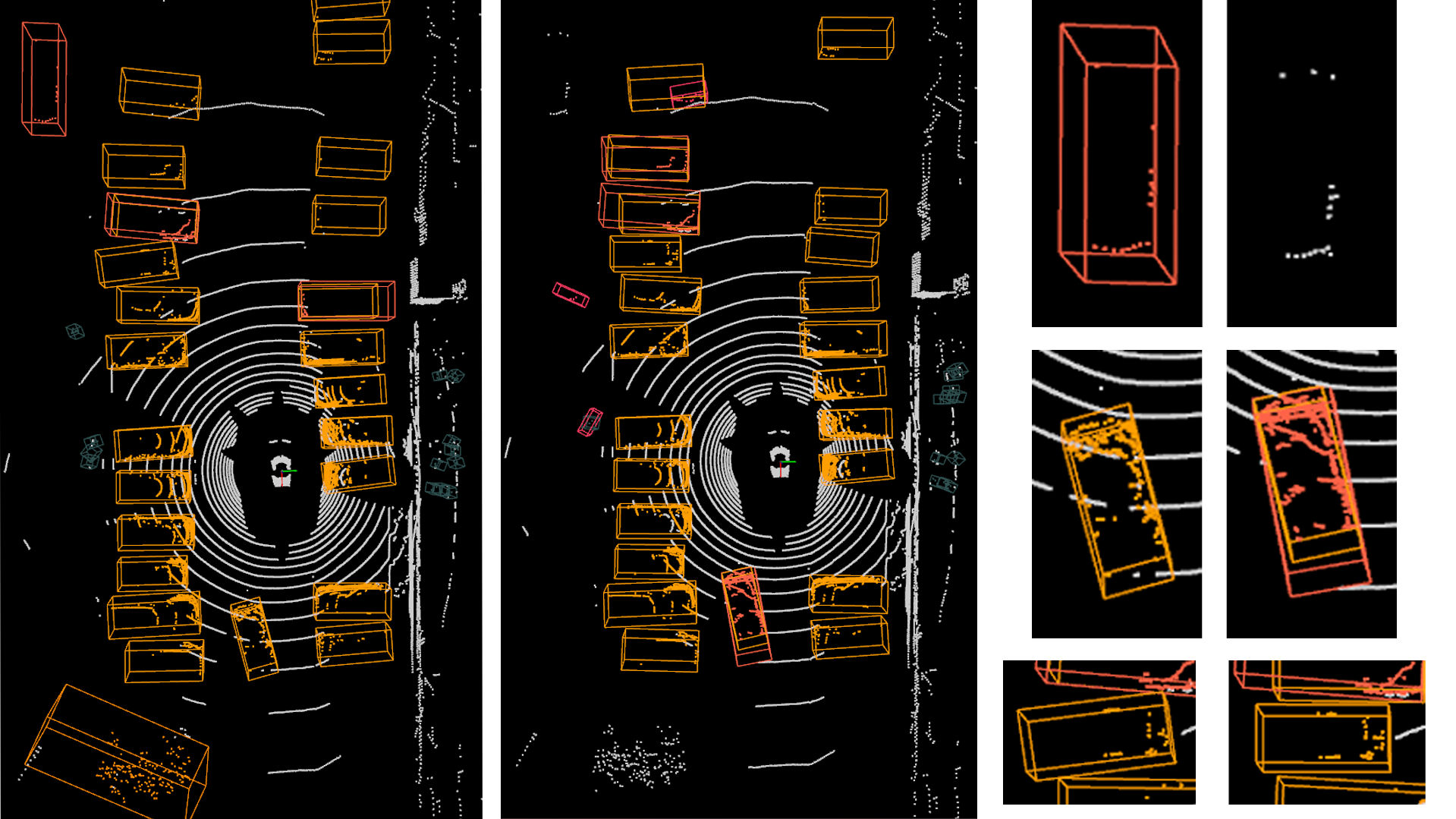}
    \caption{Two consecutive Lidar frames from the NuScenes dataset with predictions generated by a Centerpoint model with an mAP of $56.54\%$ (left) and zoomed in, hand-picked examples of detection inconsistencies between the frames: a missing detection (top-right pair), several plausible bounding boxes for a single object (center-right pair) and inconsistent, yet plausible bounding box predictions between two frames (bottom-right pair).}
    \label{fig:jittering_example}
\end{figure}

The first stage of the perception pipeline - the object detector - already introduces inaccuracies to the perception stack: estimated position, orientation, or extent of objects are predicted according to maximum likelihood within individual scans, but change and fluctuate between consecutive frames \cite{10.1007/978-3-031-72973-7_12}. The inter-frame irregularities are commonly referred to as jittering and arise due to inherently noisy sensor data, non-maxima suppression with nearly equally confident predictions and the general ambiguity of bounding boxes (see Fig.~\ref{fig:jittering_example}).  Jittering is further exacerbated when the training dataset and the deployment domain are mismatched as domain-specific datasets are often not large enough to finetune, much less train a robust object detector. The unstable predictions are passed down to the tracker, which infers non-zero velocities from the apparent position change, and the prediction module returns trajectories that intersect the ego path even if the obstacle is static -- prompting unnecessary stops despite the obstacles being properly parked. Even with reasonable velocity thresholds, the tracker frequently misidentifies static obstacles as dynamic.

While the problem of instable bounding boxes is common to most object detectors it is not reflected in the standard metrics for object detection such as mean Average Precision (mAP) and the nuScenes Detection Score (NDS). Autonomous driving software stacks can tolerate instable bounding boxes to a certain degree since Kalman filter based tracking algorithms smooth the inconsistencies. If, however, the predictions become too unstable (e.g. as a result of a lack of domain specific training data), the tracking algorithm fails to estimate realistic velocities.

We currently only have a small domain-specific dataset for finetuning our object detector available, therefore we are constrained to approaches that do not require extra training data. \cite{Feng.b} show that the explicit modeling of aleatoric uncertainty can improve convergence and detection quality of 3D object detectors under data scarcity while providing heteroscedastic uncertainty measures for individual predictions. We therefore augment CenterPoint with variance heads for position, velocity and yaw, and use these per-object positional variances to normalize apparent motion over a short observation window: A simple two-sample z-test compares velocity and uncertainty within the observation window and classifies each object as static or dynamic; for objects deemed static, we set velocities to zero and smooth positions before trajectory prediction. Our approach is practical for on-vehicle deployment since it integrates with Autoware with minimal changes and additional compute costs -- uncertainty is carried via existing covariance fields, and we reuse the tracker’s data associations.

In real-world test drives with our research vehicle EDGAR \cite{Karle.2024}, our method reduces false trajectory predictions and the resulting unnecessary stops. Because these drives lack ground truth, we benchmark our approach on nuScenes and observe parity with simple velocity thresholding (Autoware’s default). To reconcile this with field results, we inspect the joint distribution of predicted speed and Z-score: nuScenes shows two well-separated clusters -- clearly static and clearly dynamic -- so either speed or Z-score thresholds work similarly. Our road data, however, exhibits a third, intermediate cluster with the full range of apparent speeds but low–moderate Z-scores, where most false velocity estimates concentrate. This additional “jitter band" explains why Z-score filtering outperforms speed-only rules in noisier real-world settings.

Our contribution is threefold:
\begin{enumerate}
     \item A deployment-friendly motion classifier that uses detector-predicted aleatoric uncertainty and a two-sample z-test to distinguish true motion from jitter in 3D detections
    \item An integration into Autoware that reuses the tracker’s data association and standard covariance fields, requiring only a detector upgrade. The code can be found here: \href{https://github.com/TUMFTM/autoware_motion_classifier}{https://github.com/TUMFTM/autoware\_motion\_classifier}
    \item An empirical study showing parity with velocity thresholding on nuScenes and substantive reductions of false dynamic predictions and corresponding planner interventions in real-world driving without additional variance calibration.
\end{enumerate}

\section{Related Work}
\subsection{Uncertainty Estimation in Object Detection}
Uncertainty in object detectors is typically decomposed into epistemic and aleatoric components. Epistemic (model) uncertainty quantifies ambiguity over the parameter values that best explain the observed data, whereas aleatoric (data) uncertainty captures noise intrinsic to the sensing process and environment \cite{Hullermeier.2021}.

\textit{Epistemic uncertainty.} Practical estimation commonly relies on approximate Bayesian inference. Deep ensembles \cite{10.5555/3295222.3295387} train multiple networks independently and treat the ensemble as a mixture at inference; the ensemble mean serves as the prediction and the dispersion as an uncertainty measure. Monte-Carlo dropout \cite{Gal.2016} uses stochastic dropout at test time to draw samples from the predictive distribution; this can been applied to both 3D LiDAR object detection \cite{Feng.} and to 2D camera detection \cite{Zhao.2024}. Both approaches require multiple forward passes and are usually not real-time capable. Multiple Input Multiple Output (MIMO) networks are a single-pass alternative which train independent subnetworks within one model \cite{Havasi.2020}, exploiting the overparametrization of detector architectures. The MIMO principle is adapted to LiDAR 3D detection by \cite{Pitropov.}.

\textit{Aleatoric uncertainty.} Aleatoric uncertainty is modeled by treating network outputs as random variables. For classification, it is represented by a categorical distribution obtained via a softmax layer \cite{DiFeng.2022}. For regression targets, direct heteroscedastic modeling augments detectors with variance heads and optimizes a likelihood-based objective, which naturally down-weights noisy observations during training while learning to estimate the appropriate uncertainty level \cite{Feng., Feng.b}
. Beyond single-variance parameterizations, mixture-density formulations employ Gaussian Mixture Models to capture more complex distributions \cite{Choi.}. LaserNet \cite{Meyer.2019} predicts bounding boxes as a mixture of Laplace distributions, aggregates per-point distributions via mean shift, and exploits variance in variance-aware NMS. \cite{Meyer.2020} further accounts for label noise by estimating aleatoric uncertainty with an explicit model of ground-truth uncertainty.

\subsection{Multi-Object Tracking and Movement Classification}
\textit{Multi-object tracking.} Model-based multi-object tracking assumes an explicit motion and measurement model (e.g., constant-velocity or bicycle) and estimates latent object states with Bayesian filters (Kalman/EKF/UKF or particle filters) as in \cite{Li., Li.2024}. Process and measurement noise are handled within the filter, yielding the maximum-a-posteriori (or minimum-variance state) given noisy detections. Because detections are unlabeled -- and may be ambiguous or multimodal -- data association is required to link measurements to tracks; standard solutions include Joint Probabilistic Data-Association \cite{BarShalom.2009}, as well as global optimization via Hungarian assignment \cite{Kuhn.1955} or min-cost flow formulations \cite{Chen.2023}. 
Data-driven approaches employ different types of neural network architectures such as convolutional neural networks \cite{Sun.2021}, graph neural networks \cite{Buchner.2022} or transformers \cite{TimMeinhardt.2022} to learn complex relationships of data association and movement from annotated sequences. \cite{Wei.2025} integrate model-based and data-driven approaches by enhancing a Bayesian estimator with neural networks where it is deemed too simplistic.

\textit{Movement classification.} Static–dynamic discrimination spans pixel, point, and track levels. For fixed cameras, layered statistical models label pixels as background, moving foreground, or static foreground, enabling stop-and-go handling and robust occlusion recovery \cite{Gallego.1012200810152008}. \cite{Omrani.2020} employ an optical-flow algorithm and motion cues to separate static from dynamic obstacles in a camera detection and tracking pipeline. For LiDAR, recent work \cite{AlipourSormoli.2024} replaces iterative closest point (ICP) pointcloud registration with an optical-flow formulation on 2.5D grids to estimate object motion state and velocities more computationally efficient using point-level data. \cite{Adnan.2024} develop a LiDAR track-level method to classify the motion state of objects after detection and tracking by inspecting ego-motion–compensated speed and heading as well as their short-window stability.
\section{Methods}
\subsection{Uncertainty Aware Object Detection}\label{sec:UAOD}
We adopt CenterPoint \cite{Yin.2021} as our 3D object detector because it is already integrated into the Autoware software stack deployed on our research vehicle.

To capture aleatoric uncertainty, we treat the bounding-box regression targets $\mathbf{x}= [x,\; y,\; v_x,\; v_y,\; \gamma_{\text{yaw}}]$ as Gaussian probability distributions that need to be modeled. The standard regression outputs $\mathbf{x}$ serve as the mean. We follow the direct modeling paradigm \cite{Kendall.2017, Feng.} and use additional heads to predict the variances of these distributions. Therefore, the regression loss function for our model becomes the Gaussian negative log likelihood loss \cite{Nix.1994} 
\begin{equation}
\mathcal{L}_{\text{GNLL}}
=\frac{1}{2}\exp(-\mathbf{x}_{\sigma})\,
\lVert \bar{\mathbf{x}}-\mathbf{x} \rVert^{2}
+\frac{1}{2}\mathbf{x}_{\sigma},
\end{equation}
where \(\bar{\mathbf{x}}\) denotes the ground-truth vector.
For numerical stability we estimate the log-variances $\mathbf{x}_{\sigma} = \log(\boldsymbol{\sigma}_{\mathbf{x}}^{2})$ instead of the variances themselves and clamp each element of \(\mathbf{x}_{\sigma}\) to \([-10,\infty)\). Note that uncertainties for width, length, height, and \(z\)-position of the bounding boxes are not modeled.

\begin{figure*}[h!]
    \centering
    \includegraphics[width=0.98\linewidth]{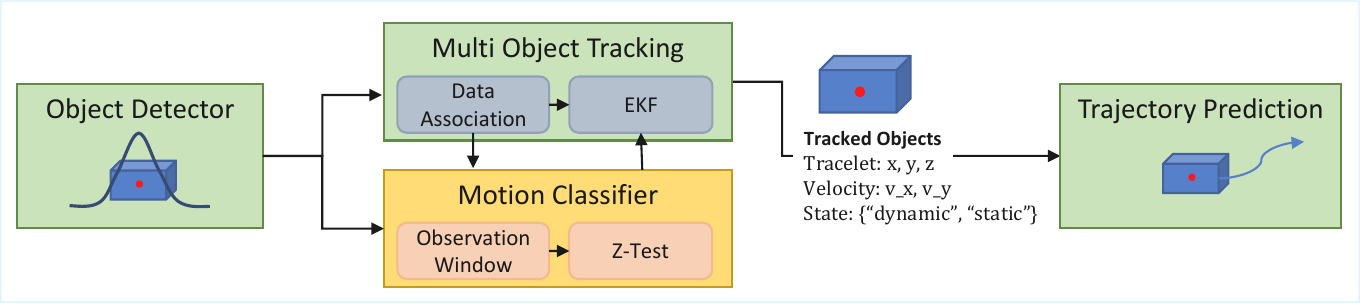}
    \caption{Integration of the motion classifier into the perception software stack of an autonomous vehicle: The motion classifier exists in parallel to the multi object tracking module. It uses the data association of the tracker and returns smoothed positions and zero-velocity if it classifies an object as static. The tracked objects are passed to the trajectory prediction algorithm without the need for an adaptation of interfaces.}
    \label{fig:integration}
\end{figure*}

\subsection{Object Motion Classifier}\label{sec:MC}
We leverage the availability of variance estimates to determine whether an observed change in position is a result of the object detector’s uncertainty, the jittering, or whether it stems from a real movement of the object. To do so, we collect several consecutive detections from each present object and divide the observation window in two. Each part of the observation window acts now as a sample. If the means of the two samples are different by more than what is explained by the noisiness of the detections, the object must have moved during the observed time frame. 

We use a two-sample location Z-test \cite{Sprinthall.2012} to decide whether the difference in the means is significant enough to classify an object as dynamic given the uncertainty of the detections. More formally, we are testing the null hypothesis $H_0$ “static object” against the alternative hypothesis $H_1$ “dynamic object”. We reject the null hypothesis if the Z-score exceeds a threshold $\alpha$:
\begin{equation}\label{eq:z_test}
    z \;=\; \frac{\lvert \bar{\mu}_{1} - \bar{\mu}_{2} \rvert}{\sqrt{\dfrac{\hat{\sigma}_{1}^{2}}{n_1} + \dfrac{\hat{\sigma}_{2}^{2}}{n_2}}} \; > \;\alpha ,
\end{equation}
with $\bar{\mu}_i, \hat{\sigma}_{i}^{2}, n_i \;\text{and}\; i\in\{1, 2\}$ as sample mean, observed variance mean and number of observations in both samples. The uncertainty aware object detector allows us to calculate the observed variance mean instead of estimating it from the sample:
\begin{equation}\label{eq:var}
    \hat{\sigma}_i^2 = \frac{1}{n_i} \sum_{j=t_i}^{t_i+n_i} \sigma_j^2\;,
\end{equation}
where $\sigma_j^{2}$ is the estimated variance for one parameter of a single bounding box in the sample starting at time $t_i$. 

The two-sample $Z$-test relies on three conditions:
\begin{enumerate}
  \item[(i)] \textit{Approximate normality of the sample means}.  
        Each detector output is modeled as normal distribution $\mathcal{N}(\mu,\sigma^{2})$; their aggregation therefore remains Gaussian.
  \item[(ii)] \textit{Known population standard deviations}.  
        The detector supplies variances for every detection, so the population standard deviation $\hat{\sigma}_i^2$ in eq. \eqref{eq:z_test} can be computed directly using eq. \eqref{eq:var} rather than estimated from the data which would require a significantly larger sample size.
  \item[(iii)] \textit{Independent observations}.  
        Successive detections of the same object are serially correlated, so treating the window lengths $n_i$  as counts of independent samples inflates the test statistic.  
        We counter this by collapsing each window into a single cluster, retaining its empirical mean and aggregated population variance but setting the effective sample size to one.  
        This adjustment brings the test statistic closer to its theoretical $Z$-distribution, yet it still requires a posterior tuning of the decision threshold.
\end{enumerate}

Finally, the observation-window length trades speed for robustness: short windows detect state changes rapidly but are noise-sensitive, whereas longer windows respond more slowly but yield more reliable estimates of the population mean and standard error.

\subsection{Integration into the Software Stack}

A key design goal of the motion classifier is compatibility with existing autonomous driving software stacks. To achieve this, the classifier is integrated in a way that minimizes interference with existing module logic and interfaces. Wherever possible, existing infrastructure -- such as messaging formats and processing pipelines -- is reused to ensure easy deployment and maintainability (Fig. \ref{fig:integration}).

The only major change required is the replacement of the object detector. Since the motion classifier relies on uncertainty estimates to assess object movement, the detector must be upgraded to produce variance estimates, as described in section \ref{sec:UAOD}. These uncertainty values need to be propagated through the pipeline. Fortunately, the standard ROS2 message type for detected objects already includes a covariance field, which allows this additional information to be carried forward without modifying the message structure or downstream modules.

To associate new detections with their corresponding past observations, the motion classifier requires data association, which is typically computationally intensive. To avoid redundant computation, the motion classifier reuses the data association results from the multi-object tracker. The tracker already performs detection-to-track matching using maximum score matching according to \cite{Wang.2019}. These associations are passed to the motion classifier, allowing it to either update existing observation windows or initialize new ones. Each observation window is linked to a tracked object via its unique track ID.

Once an observation window contains a sufficient number of detections, the classifier evaluates the object's motion in $x$ and $y$ direction independently. It compares $max\{z_x, z_y\}$ to the threshold $\alpha$ and outputs a motion state: either \textit{static} or \textit{dynamic}. This motion state is incorporated into the tracked object’s representation. Note that we do not use yaw or the velocities and their variances as estimated by the CenterPoint detector. While it is necessary to incorporate them during training in order to keep a consistent loss for all regression targets, including them into the motion classifier has a detrimental effect to its performance due to poor calibration of their variance estimates.

In the Autoware stack, the trajectory prediction module primarily reacts to an object’s estimated velocity. Therefore, for objects classified as static, the velocity components are set to zero in the tracked object's state model. To further mitigate positional jittering, the object's position is replaced with the average position over the observation window. Although this positional smoothing is not strictly necessary (as zero velocity suppresses prediction), it helps maintain spatial consistency.

Objects classified as dynamic remain unaffected and are handled entirely by the multi-object tracker. Ultimately, all tracked objects -- augmented with motion state and smoothed positions as needed -- are forwarded to the trajectory prediction module.

\section{Experiments}

\subsection{Uncertainty Aware Object Detector}

We train two CenterPoint variants  --  each with and without variance heads $\sigma²_x$, $\sigma²_y$, $\sigma²_{v_x}$, $\sigma²_{v_y}$ and $\sigma²_\theta$ (however we drop $\sigma²_{v_x}$, $\sigma²_{v_y}$, $\sigma²_\theta$ after training). For nuScenes-only experiments we switch to a VoxelNet encoder (voxel size 0.075~m × 0.075~m × 0.2~m), matching the best-performing configuration \cite{mmdet3d2020}. We train for 30 epochs on all ten nuScenes classes, again comparing models with and without the uncertainty heads.

For on-vehicle use and real-world experiments we adopt a lightweight PointPillars encoder and change the point-cloud range from 51.2~m (the standard nuScenes setting) to 100~m. We pretrain 20 epochs on nuScenes (car, truck, bus, bicycle, pedestrian), reset the learning-rate schedule, and fine-tune 20 epochs for domain adaptation on our own dataset using a 100~m point-cloud range. Our domain-specific data comprises 1,357 samples in 15 scenes for training, 320 frames in 4 scenes for validation, and 283 samples in 3 scenes as test split. The dataset covers about 16 minutes of driving with professional annotations.

Table \ref{tab:ODs} compares the performance of the uncertainty-aware CenterPoint models with their vanilla baseline. On nuScenes only, the uncertainty heads slightly reduce mAP (0.556 vs. 0.591). After pretraining on nuScenes and finetuning on our data, the uncertainty-aware model improves markedly (0.478 vs. 0.375) over the vanilla baseline. Most of the gain comes from rarer classes (truck, bus, bicycle, pedestrian). This suggests that modeling aleatoric uncertainty helps learning the domain shift from a relatively small amount of data by down-weighting noisy or occluded samples and focusing learning on informative ones, which is in line with \cite{Feng.b}. Note that cross-regime comparisons are not meaningful due to differing backbones, classes, sensor setups and environments.

\begin{table}[pbth]
\vspace*{3mm}
\caption{Performance of our uncertainty aware object detectors compared with their baseline. Real-world experiments are conducted with the uncertainty-aware detector pretrained on nuScenes and fine-tuned on a small domain specific dataset.}
\centering
\begin{tabular}{llcc}
\toprule
Dataset & Network & mAP & pos. ECE\\
\midrule
\multirow{2}{*}{nuScenes only}    & Vanilla            & 0.591 & - \\
                             & Uncertainty-aware  & 0.556 & 0.056\\[0.3em] 
\cline{1-4}\\[-0.2em] 
\multirow{2}{*}{\makecell[l]{nuScenes +\\domain specific dataset}} & Vanilla            & 0.375 & - \\
                             & Uncertainty-aware  & 0.478 & 0.138\\
\bottomrule
\end{tabular}
\label{tab:ODs}
\end{table}

\begin{figure}[!htbp]
\centering
\vspace{0.5em}
\captionsetup[sub]{skip=10pt} 
\hspace*{0.04\linewidth}
\begin{subfigure}[t]{0.44\linewidth}
    \centering
    \fontsize{8pt}{10pt}\selectfont
    \def\svgwidth{\linewidth}
    \input{nuScenes_var_calibration.pdf_tex}
    \caption{nuScenes only}
\end{subfigure}\hspace{0.03\linewidth}
\begin{subfigure}[t]{0.44\linewidth}
    \centering
    \fontsize{8pt}{10pt}\selectfont
    \def\svgwidth{\linewidth}
    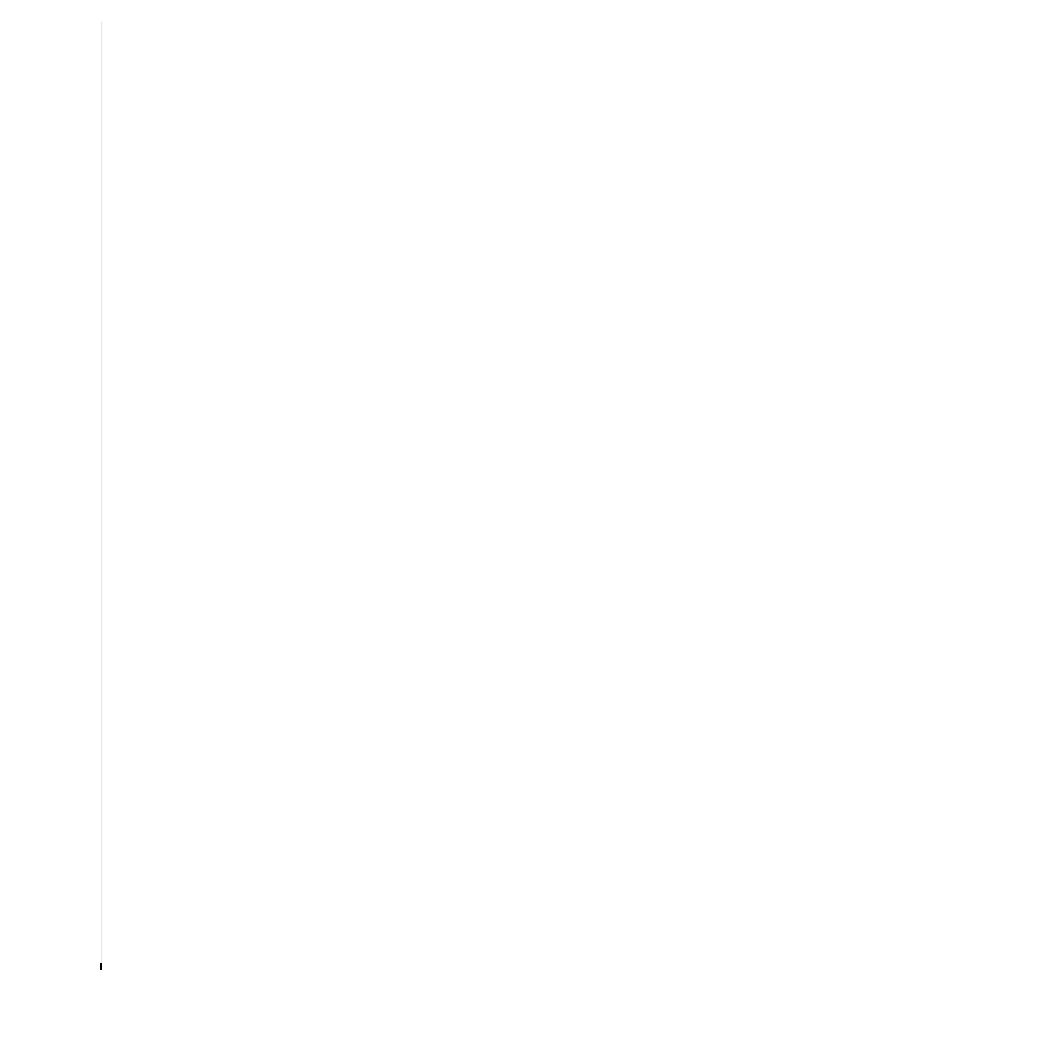
    \caption{Our data}
\end{subfigure}
\caption{Calibration of predicted vs. observed positional uncertainty. The plots show the standard deviation (SD) of the observed translational error in $x$ and $y$, binned by predicted SD. On nuScenes, the uncertainty-aware detector shows strong calibration. With finetuning on our smaller dataset and longer detection range, the correlation remains but is weaker: predicted SD in $x$ tends to be overestimated, while in $y$ it is consistently underestimated.}
\label{fig:calib}
\end{figure}

\begin{figure*}[!htb]
  \centering
    \vspace{0.5em}
  \begin{subfigure}[t]{0.9\textwidth}
    \centering
    \includegraphics[width=\textwidth]{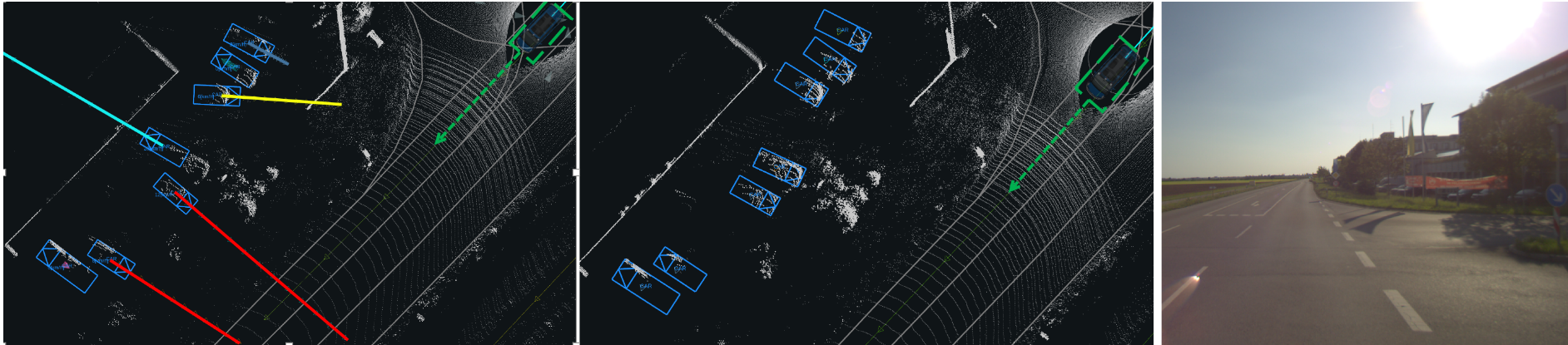}
    \caption{Predicted velocities (from top to bottom, length of trajectory corresponds to predicted velocities): 0.6 m/s (blue), 1.7 m/s (yellow), 7.8 m/s (light blue), 3.3 m/s (red), 2.2 m/s (red).}
    \label{fig:vert-a}
  \end{subfigure}

  \vspace{0.6em}

  \begin{subfigure}[t]{0.9\textwidth}
    \centering
    \includegraphics[width=\textwidth]{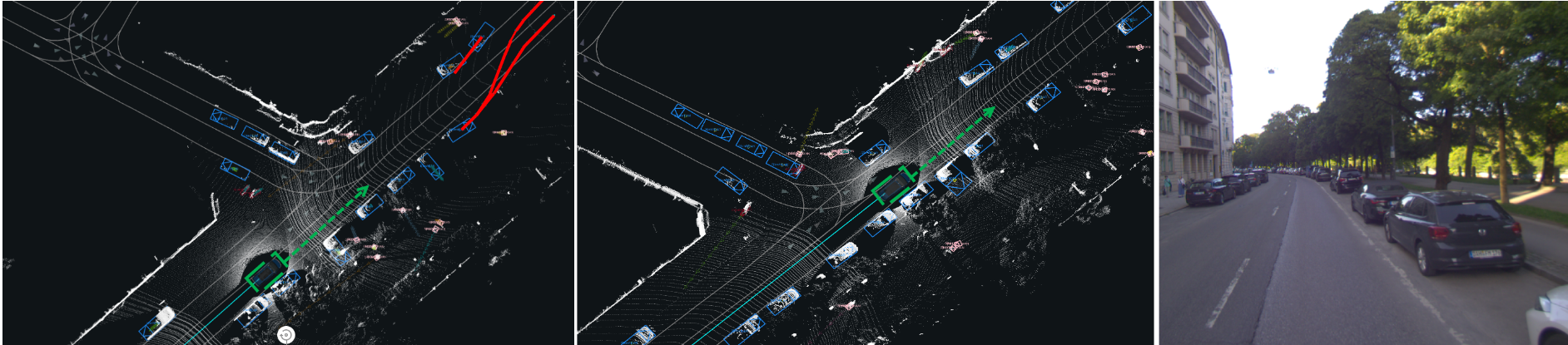}
    \caption{Predicted velocities (from top to bottom, length of trajectory corresponds to predicted velocities): 0.6 m/s against the direction of driving (red), 5.0 m/s merging into lane (red)}
    \label{fig:vert-b}
  \end{subfigure}

  \caption{Critical scenes where jittering makes static vehicles appear to move, producing false trajectories that conflict with the ego path (conflicting trajectories in red, ego vehicle and direction of motion in green). Left: baseline predictions. Middle: same scenes with our jittering-mitigation module enabled, suppressing the spurious trajectories. Right: corresponding front-camera view.}
  \label{fig:failure}
\end{figure*}

Table~\ref{tab:ODs} also reports the mean expected calibration error (ECE) for the estimated positional variances, $\sigma_x^2$ and $\sigma_y^2$. On our dataset we observe more pronounced miscalibration. This is not solely due to the data itself, but can be partly attributed to detection range. Evaluating the nuScenes model used to pretrain our deployed model -- which was trained for a 100~m detection range, a nearly twofold increase to the nuScenes standard range -- we obtain a positional ECE of 0.102. This suggests that roughly half of the increase arises from training with a larger point-cloud range, which introduces sparser, more occluded objects and therefore a harder task. Fig. \ref{fig:calib} shows calibration plots for both uncertainty-aware object detectors, binning variance predictions and linking them to the variance of the observed error. To gain a better intuition about the positional errors, variances are converted to the standard deviation (SD). The nuScenes-only detector achieves excellent calibration, as indicated by the ECE of 0.056. On the high end of predicted variances, where only relatively few predictions are found, it mostly underestimates the error. Our deployed detector shows more pronounced miscalibration; interestingly, it overestimates variance longitudinally for mid-to-high predictions while consistently underestimating it laterally. We hypothesize that, beyond the longer detection range, changes in noise characteristics and data distribution -- together with the size mismatch between the large pretraining dataset and the much smaller finetuning dataset -- underlie this shift in calibration.

\subsection{Motion Classifier}

\begin{figure*}[htbp]
\centering
\begin{subfigure}[t]{0.32\textwidth}
    \centering
    \includegraphics[width=0.99\linewidth]{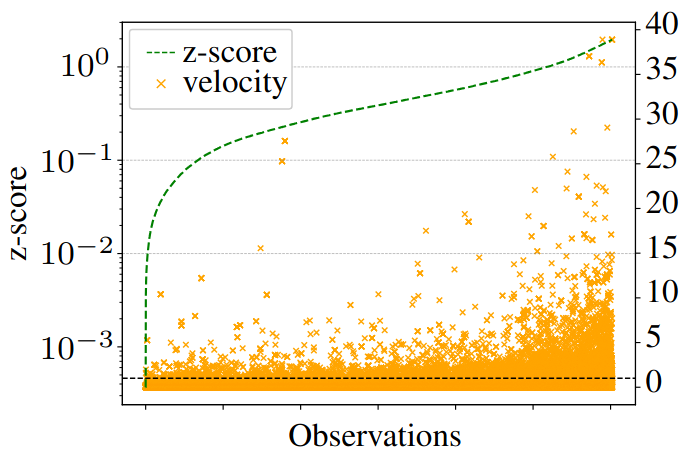}
    \caption{Vehicles}
\end{subfigure}
\begin{subfigure}[t]{0.32\textwidth}
    \centering
    \includegraphics[width=0.99\linewidth]{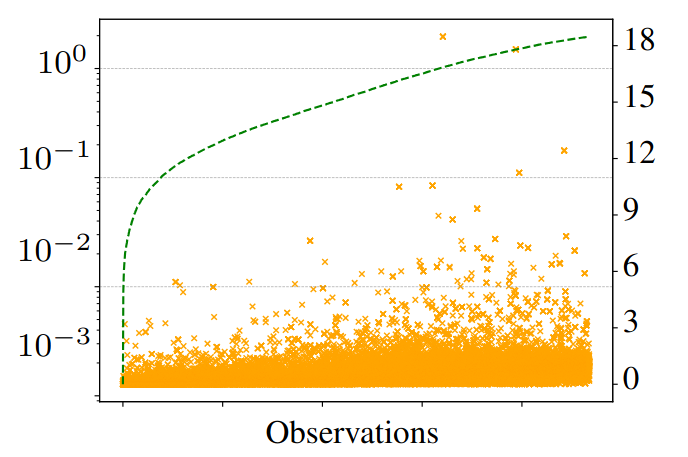}
    \caption{Pedestrians}
\end{subfigure}
\begin{subfigure}[t]{0.32\textwidth}
    \centering
    \includegraphics[width=0.99\linewidth]{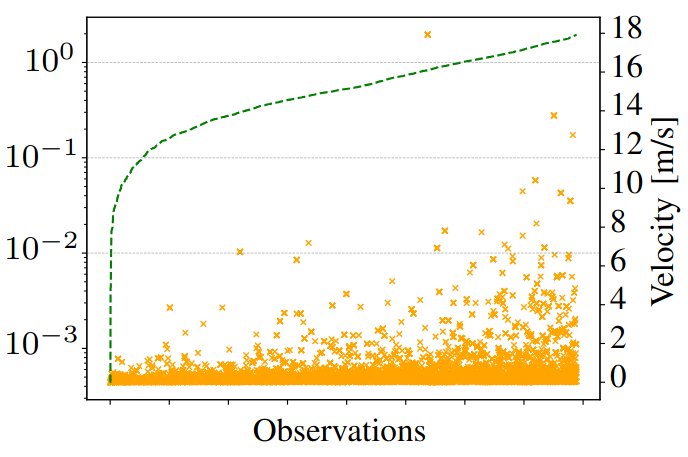}
    \caption{Bicycles}
\end{subfigure}
\caption{Z-scores and predicted velocities for detections classified as static (Z-score below the threshold). Z-score filtering removes many unrealistic velocity predictions compared to a simple velocity threshold. For vehicles, the black dashed line marks the Autoware velocity cut-off; predictions below this value are suppressed. No such cut-off exists for the other classes.}
\label{fig:z_vel}
\end{figure*}

Fig. \ref{fig:failure} highlights two recurring failure cases from our road tests. In \ref{fig:vert-a}, our research vehicle passes diagonally parked cars. As we approach, the vehicles mutually occlude one another; the changing occlusion shifts bounding boxes, causing jitter and spurious velocity estimates for static cars. Because their headings are roughly perpendicular to the ego heading, these false motions often intersect the planned path and trigger unnecessary stops. The problem is amplified by additional clutter/occlusion (e.g., trees, fencing) which increases localization uncertainty. Figure \ref{fig:vert-b} shows a driving scenario with parallel-parked cars along the curb. At longer ranges detections jitter noticeably; when interpreted as velocity, they are falsely predicted into our lane -- sometimes even against the direction of driving. The planner then (appropriately) slows or stops, despite the objects being parked in reality. Fig. \ref{fig:failure} shows that our approach is able to reduce these errors in real world scenarios by comparing the apparent movement of the bounding boxes with the estimated aleatoric uncertainties.

Fig. \ref{fig:z_vel} plots predicted velocities for objects whose Z-scores fall below the dynamic threshold -- i.e., cases where motion predictions are suppressed. Although we initially intended applying the Z-score filter only to vehicles, in practice it is also able to suppress false velocity predictions for pedestrians and bicycles. As shown, all classes exhibit unrealistic velocity estimates due to bounding box jittering that the filter now discards.

Initially, we set the threshold to $\alpha=1.96$, which would correspond to a 5\% probability of the two-sided Type I error (wrongly classifying an object as dynamic) under the z-test assumptions. As explained in sec. \ref{sec:MC} the assumptions are not completely fulfilled in our use case; therefore $\alpha$ requires manual tuning. Unfortunately, we lack the ground-truth data to quantitatively assess the choice of $\alpha$, but our results on nuScenes show that the optimal threshold might be slightly higher than 1.96. The Type II error (classifying a dynamic object as static) cannot be fixed a priori because it depends on the unknown true speed distribution of other agents. While Type I errors largely revert to Autoware’s default behavior (predicting motion for static objects) and therefore are not worse than the baseline, Type II errors may introduce new failure modes in which dynamic objects are assigned a velocity of zero, even though their positions continue to be updated. Empirically, we do not observe a notable increase in poor decisions due to missing velocity predictions. Likely reasons are: (i) truly moving objects tend to be well separated and unobstructed, yielding lower aleatoric uncertainty, less jitter, smaller predicted variances, and thus are more often labeled dynamic due to higher Z-scores; and (ii) even when labeled static, object positions continue to update. As a result, the planner accounts for their changing location but without projecting their motion into the future.

In practice, we need the classifier to come to a quick first decision. Therefore, we use a minimal observation window size of two. We acknowledge that the classifier is more error-prone during the first few observations, before the observation window is sufficiently populated. As detections accumulate over a sliding window of ten cycles, the decision stabilizes and aligns increasingly with the safety driver’s assessment.

\subsection{Validation of our Approach}

We lack ground-truth motion labels for our road tests, so we validate the approach on the nuScenes dataset. Because nuScenes does not support resimulation through Autoware, we build an offline pipeline: First, we run a CenterPoint detector (see Tab. \ref{tab:ODs}), then use PolyMOT \cite{Li.} as tracker and data associator for creating the observation windows, which we pass to our motion classifier. We evaluate the predicted motion states against nuScenes ground truths. nuScenes provides movement related per-object attributes (e.g., moving, stopped, parked) exclusively for the vehicle and pedestrian classes. We map these attributes to “dynamic” or “static" , enabling a direct, label-based assessment of our classifier.

We compare our method to simple velocity thresholding -- the heuristic used in the original Autoware software stack -- using Average Precision (AP) for vehicles and pedestrians (Tab. \ref{tab:zscore_vs_velocity}). On the nuScenes validation split, our approach performs on par with the baseline. In real-world runs, however, it prevents many false dynamic predictions that would otherwise cause planner slowdowns, disengagements, and safety-driver interventions. 

\begin{figure}[htbp]
\centering

\vspace{0.5em}

\begin{subfigure}[t]{0.45\textwidth}
    \centering
    \includegraphics[width=0.99\linewidth]{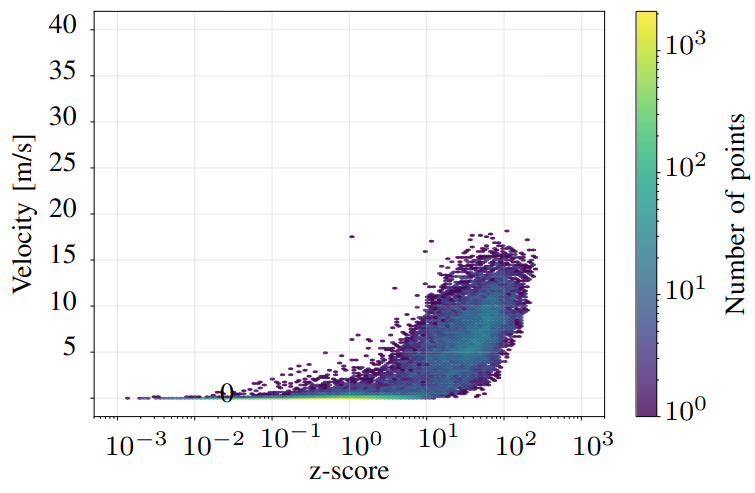}
    \captionsetup{skip=9pt}
    \caption{nuScenes}
\end{subfigure}

\vspace{0.6em}

\begin{subfigure}[t]{0.45\textwidth}
    \centering
    \includegraphics[width=0.99\linewidth]{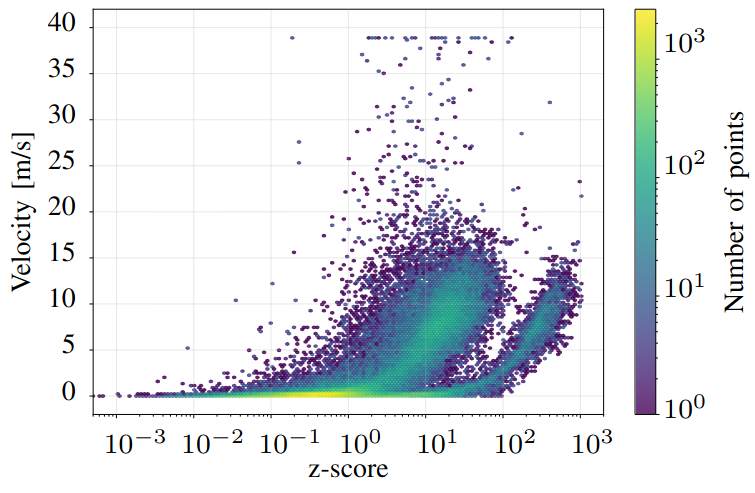}
    \captionsetup{skip=9pt}
    \caption{Real-world test drive}
\end{subfigure}

\caption{Density of predicted speed vs. Z-score for vehicles. (a) nuScenes: two well-separated modes: near-zero speed with small Z-scores (static) and typical urban speeds with large Z-scores (moving). Velocity-thresholding and Z-score filtering make similar decisions. (b) Real-world test drive: an additional “jitter band” with non-zero speed but low–moderate Z-scores; speed-only thresholds mislabel this band as dynamic, whereas a Z-score cutoff suppresses it better.}
\label{fig:nu_vs_ed}
\end{figure}

To explain this discrepancy, we qualitatively analyze the distributions of Z-scores and predicted velocities on nuScenes versus our road data. Fig. \ref{fig:nu_vs_ed} shows densities of predicted speed versus Z-score; large Z-scores can come from either large velocity or low estimated variance. On nuScenes, two modes are cleanly separated: (i) near-zero speed with small $z$ (static) and (ii) typical urban speeds with large $z$ (moving). Because these modes are separable along both axes, a simple speed cutoff and a Z-score cutoff produce nearly identical decisions, matching the parity we observe with velocity thresholding in Tab. \ref{tab:zscore_vs_velocity}.
\begin{table}[htb]
\centering
\caption{AP results for Z-score filtering vs. simple velocity thresholding show parity of both approaches on nuScenes data.}
\label{tab:zscore_vs_velocity}
\begin{tabular}{lcc}
\toprule
Method & Vehicles (AP) & Pedestrians (AP) \\
\midrule
Z-score filtering & 0.94 & 0.98 \\
Velocity thresholding & 0.95 & 0.98 \\
\bottomrule
\end{tabular}
\end{table}
In our road data, a third regime emerges between those two modes: a jitter band with non-zero speed but low--to--moderate Z-scores. Points in this band arise when apparent motion is dominated by bounding-box wobble and transient association errors; equivalently, the inferred velocity is small relative to its uncertainty. Ideally, this band would occupy the same $z$ range as clearly static objects, but Z-scores can drift upward if uncertainty is underestimated for jittering objects or if small true motion is mixed with jitter within the observation window. As analyzed in Section~\ref{sec:UAOD}, our detector underestimates variance laterally and tends to overestimates it longitudinally. This contrasting behavior broadens the $z$-score distribution, potentially overlapping the lower $z$-range of truly dynamic objects. Although we did not observe dynamic objects being misclassified as static during our test drives, the increased miscalibration of variance estimates in our data contravenes the Z-test’s assumption of known population variances, making its application theoretically less rigorous. We plan to address this calibration issue as more data from our system and operating domain become available.

The different structure of predicted velocity and Z-score explains the discrepancy between nuScenes results and our on-road experience. Any reasonable velocity threshold classifies the jitter band as “dynamic,” yielding false trajectory predictions and planner slowdowns. In contrast, a Z-score threshold rejects much of this band and primarily counts the high-$z$ mode as dynamic, which reduces unnecessary interventions on-road. 

\section{Conclusion}

Our study shows that a simple, deployment-friendly motion classifier based on uncertainty-aware object detection and statistical testing mitigates a common real-world failure. It is able to suppress jitter in 3D detections of static objects which otherwise cascades into false trajectory predictions and unnecessary planner interventions. 
Empirically, the classifier reduces spurious stopping in road tests and matches the accuracy of simple speed thresholding on nuScenes. A density analysis of predicted speed versus Z-score reconciles these findings: nuScenes exhibits two clean modes (clearly static vs. clearly dynamic), while our road data reveals an intermediate “jitter band” with non-zero speeds but low–moderate Z-scores. Speed-only rules mislabel this band as dynamic, whereas our uncertainty-normalized test suppresses it. In parallel, adopting an uncertainty-aware CenterPoint improves object detection after domain adaptation, supporting the broader thesis that learning and using aleatoric uncertainty benefits downstream decisions in noisy settings.

\section*{Acknowledgment}
C. Schröder, as the first author,
devised the essential concepts of the proposed motion classifier, integrated the classifier in the autoware software stack, conducted the analysis and wrote the article.
Ž. Marcinkus adapted the CenterPoint object detector for aleatoric uncertainty estimation and implemented a prototype of the motion classifier. 
M. Lienkamp made an essential contribution to the concept of the research project. He revised the paper critically for important intellectual content. M. Lienkamp gives final approval for the version to be published and agrees to all aspects of the work. As a guarantor, he accepts responsibility for the overall integrity of the paper.

\bibliographystyle{IEEEtran}
\bibliography{Jittering_File}
\end{document}